\documentclass[journal, comsoc, 10pt]{IEEEtran}

\usepackage{cite}
\usepackage{amsmath,amssymb,amsfonts}
\usepackage{algorithm, algpseudocode}
\usepackage{graphicx}
\usepackage{textcomp}
\usepackage[dvipsnames]{xcolor}
\usepackage{commath}
\usepackage{siunitx}
\usepackage{adjustbox}
\usepackage{subcaption}
\usepackage{multirow}

\usepackage{soul,color}
\soulregister\cite7
\soulregister\ref7
\soulregister\pageref7

\DeclareSIUnit\db{dB}
\DeclareSIUnit\dbm{dBm}

\usepackage{cuted}

\usepackage{acronym}
\acrodef{kaf}[KAF]{kernel adaptive filtering}
\acrodef{klms}[KLMS]{kernel least mean squares}
\acrodef{krls}[KRLS]{kernel recurusive least squares}
\acrodef{kapa}[KAPA]{kernel affine projection algorithm}
\acrodef{lms}[LMS]{least mean squares}
\acrodef{rls}[RLS]{recursive least squares}
\acrodef{ald}[ALD]{approximate linear dependency}
\acrodef{ls}[LS]{least squares}
\acrodef{im2lms}[IM2LMS]{second-order intermodulation least mean squares}

\acrodef{rx}[Rx]{receive}
\acrodef{tx}[Tx]{transmit}
\acrodef{rf}[RF]{radio frequency}
\acrodef{dac}[DAC]{digital-to-analog converter}
\acrodef{adc}[ADC]{analog-to-digital converter}
\acrodef{pa}[PA]{power amplifier}
\acrodef{snr}[SNR]{signal-to-noise ratio}
\acrodef{lna}[LNA]{low noise amplifier}
\acrodef{lo}[LO]{local oscillator}
\acrodef{fdd}[FDD]{frequency division duplex}
\acrodef{csf}[CSF]{channel select filter}
\acrodef{ca}[CA]{carrier aggregation}
\acrodef{osf}[OSF]{oversampling factor}
\acrodef{fir}[FIR]{finite impulse response}
\acrodef{dc}[DC]{direct-current}
\acrodef{imd2}[IMD2]{second-order intermodulation distortion}
\acrodef{rb}[RB]{resource block}
\acrodef{i}[I]{in-phase}
\acrodef{q}[Q]{quadrature}
\acrodef{lte}[LTE]{long term evolution}

\acrodef{tdl}[TDL]{tapped delay line}

\acrodef{nmse}[NMSE]{normalized mean squared error}
\acrodef{snr}[SNR]{signal-to-noise ratio}
\acrodef{sinr}[SINR]{signal-to-interference-and-noise ratio}
\acrodef{icn}[ICN]{interference-to-noise ratio}

\acrodef{svm}[SVM]{support vector machine}

\acrodef{cpd}[CPD]{canonical polyadic decomposition}

\usepackage[hyphens]{url}
\usepackage[hidelinks]{hyperref}
\hypersetup{breaklinks=true}
\def\BibTeX{{\rm B\kern-.05em{\sc i\kern-.025em b}\kern-.08em
		T\kern-.1667em\lower.7ex\hbox{E}\kern-.125emX}}

\begin{document}

\title{Complex-valued Adaptive System Identification via Low-Rank Tensor Decomposition}

\author{Oliver Ploder,
Christina Auer,
Oliver Lang, 
Thomas Paireder and
Mario Huemer
\thanks{Acknowledgments: The financial support by the Austrian Federal Ministry for Digital and Economic
Affairs, the National Foundation for Research, Technology and Development and the Christian Doppler
Research Association is gratefully acknowledged.}
\thanks{O. Ploder, C. Auer and T. Paireder are with the Christian Doppler Laboratory for Digitally Assisted RF Transceivers for Future Mobile Communications, Institute of Signal Processing, Johannes Kepler University, Linz, Austria (e-mail: oliver.ploder@jku.at).}
\thanks{O. Lang and M. Huemer are with the Institute of Signal Processing, Johannes Kepler University, Linz, Austria}}

\maketitle

\begin{abstract}
	Machine learning (ML) and tensor-based methods have been of significant interest for the scientific community for the last few decades. In a previous work we presented a novel tensor-based system identification framework to ease the computational burden of tensor-only architectures while still being able to achieve exceptionally good performance. However, the derived approach only allows to process real-valued problems and is therefore not directly applicable on a wide range of signal processing and communications problems, which often deal with complex-valued systems. In this work we therefore derive two new architectures to allow the processing of complex-valued signals, and show that these extensions are able to surpass the trivial, complex-valued extension of the original architecture in terms of performance, while only requiring a slight overhead in computational resources to allow for complex-valued operations. 
\end{abstract}

\begin{IEEEkeywords}
	LMS, low rank approximation, machine learning, tensor decomposition, Wirtinger Calculus.
\end{IEEEkeywords}

\section{Introduction}

\IEEEPARstart{D}{eep learning} (DL) and neural networks (NNs)~\cite{Goodfellow-et-al-2016} are among the most popular techniques of machine learning (ML), and are broadly used in signal processing, among other disciplines~\cite{liu2017survey,5406124,ploderVTC2019,ploderAsilomar2019}. However, the umbrella of ML covers many other techniques, such as support vector machines~\cite{auerVTC2020,Smola2004}, kernel adaptive filters~\cite{txhamchristina,liu_principe_haykin_kernel_adaptive_filtering}, random forests~\cite{hastie01statisticallearning,8827054} and tensor-based estimators~\cite{cichocki_tensor_2015}. Although it has been shown that tensor-based methods can deliver on par, or even better performance than other methods~\cite{sidiropoulos_tensor_2017,kargas_nonlinear_2019}, and can be used in a variety of applications~\cite{bousse2017tensor,favier2008tensor,6190040,fernandes2009blind,kibangou2007blind,cichocki_tensor_2015,10.1007/978-3-642-33460-3_39}, they are usually disregarded due to their high memory- and computational footprint needed to approximate a given system.

In an attempt to reduce complexity of tensor-based methods, we recently introduced~\cite{TensorPaper} the combination of tensors with least mean squares (LMS) filters for system identification with minimal model knowledge, so called Wiener and Hammerstein models~\cite{wiener_1958} or combinations thereof. We analyzed several of these combinations and came to the conclusion that the proposed methods cannot only outperform (or be on par with) architectures utilizing a single tensor or spline adaptive filters (SAFs), but significantly reduce complexity compared to these methods. However, a downside of the proposed architectures is, that they are only able to deal with real-valued input and output signals. Therefore, they are not suitable for a variety of signal processing and communications related problems. 

In this work, we extend the theory of the tensor-LMS (TLMS) block, originally presented in~\cite{TensorPaper} to allow complex-valued input and output signals. The presented theory can be trivially extended to all other architectures presented in~\cite{TensorPaper} and hence is not repeated in this work. As will be shown, the resulting architectures, still keeping complexity at an absolute minimum, yield very good performance for the simulated scenario.

\begin{figure}[t!]
	\centering
	\includegraphics[width=0.65\columnwidth]{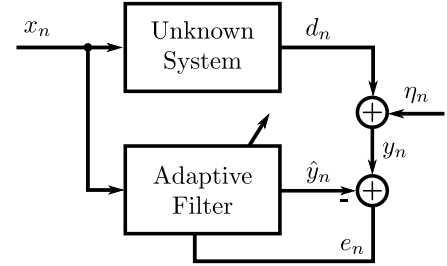}
	\caption{An overview of the general problem of adaptive system identification~\cite{TensorPaper}.}
	\label{fig:Arch}
	\vspace{-0.5cm}
\end{figure}

\section{Preliminaries and Notation}

Before reviewing the findings of~\cite{TensorPaper} and presenting our extensions, this section briefly repeats the overall problem statement and mathematical notation used in~\cite{TensorPaper}, as it is also needed in the remainder of this work.

\subsection{Preliminaries}

Like the overall problem discussed in~\cite{TensorPaper} (see Fig.~\ref{fig:Arch}), the aim is to approximate an unknown system with an adaptive filter by utilizing the same input signal $x_n$ and only observing the output $y_n$. Naturally, the ideal output of the unknown system $d_n$ is subject to noise $\eta_n$ to yield the overall output $y_n = d_n + \eta_n$. Besides updating the approximation of the system with each observed sample (i.e. one optimization step per time-step), the adaptive filter further assumes that the unknown system itself may not remain static over time~\cite{TensorPaper}.

\subsection{Tensor Background}
In this work, we adhere to the widely adopted definition of the term tensor as presented in~\cite{sidiropoulos_tensor_2017,kargas_nonlinear_2019}. That is, a tensor may be represented as an $M$-dimensional array, indexed by $i_1,i_2,i_3, \dots, i_M$~\cite{TensorPaper}. We denote a tensor by $\mathcal X$, and, like in~\cite{TensorPaper}, we use the notations $\circledcirc, \circledast, \odot$ to refer to the outer (tensor) product, the Hadamard product and the Khatri-Rao product, respectively. 

A rank-1 tensor $\dot{\mathcal X}$ of order $M$ (also called $M$-way tensor), is the outer product of a collection of $M$ vectors $\hat{\mathbf a}_m\in \mathbb R ^{I_m \times 1}$, $\forall m \in \{1,\dots,M\}$~\cite{TensorPaper}
\begin{align}
	\dot{\mathcal X} = \hat{\mathbf a}_1 \circledcirc \dots \circledcirc \hat{\mathbf a}_M 
\end{align}
%
which can also be written as~\cite{TensorPaper}
\begin{align}
	\dot{ \mathcal X} (i_1,i_2,\dots,i_M) = \hat{\mathbf a}_1(i_1) \cdots \hat{\mathbf a}_M(i_M) = \prod_{m = 1}^M \mathbf A_m(i_m,1)
\end{align}
with $\hat{\mathbf a}_{m}(i_m)= \mathbf A_m(i_m,1)$.
%
Further, any $M$-way tensor $\mathcal X $ with a higher rank than one can be decomposed into a sum of rank-1 tensors~\cite{TensorPaper}
\begin{align}
	&\mathcal{X}   =  \sum_{r = 1}^R \dot{ \mathcal{X}}^{(r)} =  \sum_{r = 1}^R \hat{\mathbf a}_{1}^{(r)} \circledcirc \dots \circledcirc \hat{\mathbf{a}}_{M}^{(r)} 
	\Longleftrightarrow \\
	&\mathcal X (i_1,i_2,\dots,i_M) =  \sum_{r = 1}^R \prod_{m = 1}^M \mathbf A_m(i_m,r)
	\label{eq:forwardtensor}
\end{align}
Additionally, the Hadamard product over all matrices $\mathbf A_{m}$ with $m \neq m'$ is defined as~\cite{TensorPaper}
\begin{align}
	\circledast_{m\neq m'} \mathbf A_{m}: = \mathbf A_M \circledast \cdots \circledast \mathbf A_{m'+1} \circledast \mathbf A_{m'-1} \circledast \cdots \circledast \mathbf A_1 \,.
\end{align}

The discretization used to obtain an index $i$ for the tensor input is given by the function~\cite{TensorPaper}
\begin{align}
	\text{disc}(x_n) &= \left\lfloor\frac{x_n}{\Delta x}\right\rfloor + \frac{N_{\text{Bins}} }{2}\label{eq:disc}
\end{align}
if $N_{\text{Bins}}$ is even and with $\Delta x$ denoting the discretization interval.	 

Lastly, the superscripts $\left(\cdot\right)^\text{T}$, $\left(\cdot\right)^\text{H}$, $\left(\cdot\right)^\text{*}$ denote the transpose, Hermitian transpose and conjugate, respectively.

\section{TLMS -- Review}\label{sec:sota}

\begin{figure}[t!]
	\centering
	\includegraphics[width=0.85\columnwidth]{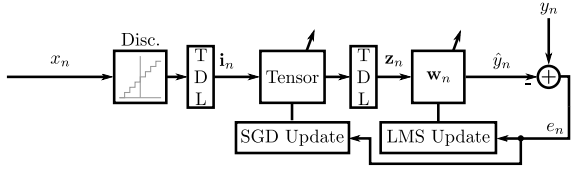}
	\caption{The original tensor-LMS architecture from~\cite{TensorPaper}, which is only able to handle real valued data.}
	\label{fig:TLMS}
	\vspace{-0.5cm}
\end{figure}

Before presenting the extensions proposed in this paper, this section reviews the TLMS approach presented in~\cite{TensorPaper} and depicted in Fig.~\ref{fig:TLMS} to introduce the most important notations. As the name TLMS implies, this adaptive filter consists of a tensor followed by an LMS filter, hence it is suitable for Hammerstein-type problems (i.e. nonlinearity before linear block). The overall output of this system is denoted by $\hat{y}_n$ and allows to express the joint cost function as~\cite{TensorPaper}
\begin{align}
	J_n &= e_n^2 =  \left(y_n - \hat{y}_n \right)^2 = \left(y_n - \mathbf{w}_{n}^\text{T} \mathbf{z}_{n} \right)^2  \\
	&= \left(y_n - \sum_{p = 1}^{P} w_{n,p} \sum_{r = 1}^R \prod_{m = 1}^M \mathbf A_{m,n-p+1}(i_{m,n-p+1},r)  \right)^2
\end{align}
where 
\begin{align}
	\mathbf z_n =& 
	\begin{pmatrix}
		z_n\\
		\vdots\\
		z_{n-P+1}
	\end{pmatrix} 
	= 
	\begin{pmatrix}
		\sum_{r = 1}^R \prod_{m = 1}^M \mathbf A_{m,n}(i_{m,n},r)\\
		\vdots\\
		\sum_{r = 1}^R \prod_{m = 1}^M \mathbf A_{m,n-P+1}(i_{m,n-P+1},r)
	\end{pmatrix}\,,
\end{align}
denotes the tapped delay line (TDL) block and where $y_n$ is the desired TLMS output~\cite{TensorPaper}.

In order to derive an update for the coefficients $\mathbf A_{m'}$ of the tensor, the gradient of the cost function is approximated by~\cite{TensorPaper}
\begin{equation}
	\mathbf G_{\text{TLMS},m',n} :=  \frac{\partial J_n}{\partial \mathbf z_n}\frac{\partial \tilde{\mathbf z}_n}{\partial \mathbf{A}_{m'}^{\text T}} \label{eq:chain_rule}
\end{equation}
with
\begin{align}
	z_{n} \approx  \sum_{r = 1}^R \mathbf A_{m'}(i_{m',n},r) \prod_{\substack{m = 1\\m\neq m'}}^M \mathbf A_{m,n}(i_{m,n},r) = \tilde z_{n}. \label{eq:x_tilde}
\end{align}
This approximation $\tilde z_{n}$ (i.e. for $\mathbf A_{m',n}$ the time is omitted) is necessary to be able to take the derivative with respect to $\mathbf A_{m'}$~\cite{TensorPaper}, as also can be seen in \cite{gebhard_diss}. 

Therefore, the tensor update is given by
\begin{eqnarray}  \label{eq:update_A}
	\mathbf A_{m',n+1}
	=
	\mathbf A_{m',n}
	+ 2 \mu_\text{Ten} e_n \mathbf S_{m',n},
\end{eqnarray}
which is evaluated for $m^\prime = 1,\ldots,M$ and with
\begin{align}
	\mathbf S_{m',n}: = \sum_{p = 1}^P w_{n,p}
	\begin{pmatrix}
		\mathbf 0_{i_{m',n-p+1}-1\times R}\\
		\circledast_{m\neq m'} \mathbf A_{m,n-p+1}(i_{m,n-p+1},:) \\
		\mathbf 0_{I_{m'}- i_{m',n-p+1} \times R}
	\end{pmatrix},\label{eq:stlmsrev}
\end{align}
where $\mathbf 0_{i_{m',n-p+1}-1\times R} \in \mathbb R^{i_{m',n-p+1}-1 \times R}$ denotes a matrix with all elements being zero~\cite{TensorPaper}.

\section{Complex-Valued TLMS}\label{sec:problem}

\begin{figure}[t!]
	\centering
	\begin{subfigure}[t]{\columnwidth}
		\centering
		\includegraphics[width=0.85\columnwidth]{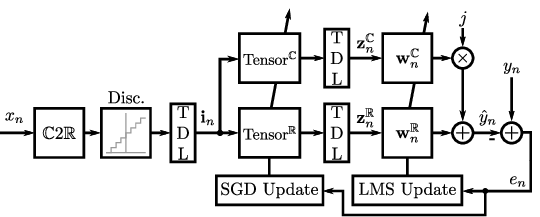}
		\caption{}
		\label{fig:model1}
	\end{subfigure}
	~ 
	\begin{subfigure}[t]{\columnwidth}
		\centering
		\includegraphics[width=0.85\columnwidth]{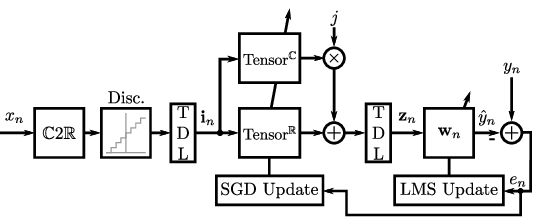}
		\caption{}
		\label{fig:model2}
	\end{subfigure}
	~ 
	\begin{subfigure}[t]{\columnwidth}
		\centering
		\includegraphics[width=0.85\columnwidth]{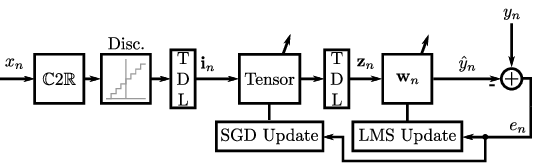}
		\caption{}
		\label{fig:model3}
	\end{subfigure}
	\caption{The proposed tensor-LMS architectures for handling complex-valued in- and output (a) via having a separate tensor and LMS for the real and imaginary paths, (b) a complex-valued LMS with two tensors, one handling the real and one the imaginary path of the model and (c) a fully complex-valued estimator.}
	\label{fig:model}
	\vspace{-0.5cm}
\end{figure}

In order to deal with complex-valued Hammerstein models (i.e. a nonlinearity followed by a linear filter) we propose three architectures based on the TLMS from~\cite{TensorPaper}. For all architectures, the input $x_n$ is the same (and complex-valued), whereby the real and imaginary parts of this signal are first stacked on top of each other to yield $\mathbf{x}_n^\mathbb{R} = \left[\text{Re}\left\{x_n\right\}, \text{Im}\left\{x_n\right\}\right]^\text{T}$ by the $\mathbb{C}2\mathbb{R}$ blocks in Fig.~\ref{fig:model}. This vector is then discretized according to~\eqref{eq:disc} and serves as an input to a two dimensional TDL (that is, two TDLs working on the rows of a matrix). The resulting output $\mathbf{i}_n$ of the TDL then serves as the input for the tensor(s). After this block, the three architectures differ in their operations, which is described in detail in the following.

Starting from the standard TLMS architecture, the first, obvious choice (denoted as TLMS-2R) is depicted in Fig.~\ref{fig:model1} and simply uses two realizations of the same architecture for the real and imaginary paths (denoted by the superscripts $\mathbb{R}$ and $\mathbb{C}$), respectively. This straightforward concept basically results in the same equations as for the simple TLMS case~\cite{TensorPaper}. However, this approach lacks the ability to utilize connections between the real and imaginary parts of the system, as they appear in complex multiplications. 

To alleviate this drawback, the second architecture (TTLMS) utilizes a complex-valued LMS (CLMS) for the linear part of the system and two tensors for the real and imaginary parts of the nonlinear part (cf. Fig.~\ref{fig:model2}). This approach enables the CLMS to leverage the interplay of real and imaginary parts of the complex signal while the update of the tensors still requires only small adaptions compared to Sec.~\ref{sec:sota}, detailed in the following. 

By re-defining the cost function as
\begin{align}
	J_n &= e_n^*e_n \\
	& =  \left(y_n - \mathbf{w}_{n}^\text{T} \mathbf{z}_{n} \right)^*\left(y_n - \mathbf{w}_{n}^\text{T} \mathbf{z}_{n} \right)\,, 
\end{align}
the update equation for the normalized CLMS becomes
\begin{align}
	\mathbf{w}_{n+1} &= \mathbf{w}_n + \mu_\text{LMS} \frac{e_n \mathbf{z}_n^*}{\epsilon + \mathbf{z}^\text{H}_n\mathbf{z}_n}\,.\label{eq:CLMSupdate}
\end{align}
The two tensors are updated via
\begin{align}
	\mathbf{A}_{m',n+1}^{\mathbb{R}/\mathbb{C}} &= \mathbf{A}_{m',n}^{\mathbb{R}/\mathbb{C}} + 2 \mu_\text{Ten}\, \text{Im}\left\{e_n \mathbf S_{m',n}^{\mathbb{R}/\mathbb{C}}\right\}\,,
\end{align}
for all $m\in[1,M]$, with
\begin{align}
	\mathbf{S}_{m',n}^{\mathbb{R}/\mathbb{C}} &= \sum_{p = 1}^P w_{n,p}^*
	\begin{pmatrix}
		\mathbf{0}_{i_{m',n-p+1}-1\times R}\\
		\circledast_{m\neq m'} \mathbf{A}_{m,n-p+1}^{\mathbb{R}/\mathbb{C}}(i_{m,n-p+1},:) \\
		\mathbf{0}_{I_{m'}- i_{m',n-p+1} \times R}
	\end{pmatrix}\,.
\end{align}
%

While this representation may reduce repetition of blocks compared to the first case, the tensors are still not able to make full use of the complex gradient. The final architecture (CTLMS) shown in Fig.~\ref{fig:model3}, reduces to (mostly) the same architecture as shown in Fig.~\ref{fig:TLMS}. The difference however is, that the input signal is split into its real and imaginary parts and the LMS as well as the tensor are now fully complex-valued in their outputs. This, of course, necessitates to derive new update equations for the tensor modeling the nonlinearity. This can be achieved by utilizing Wirtinger's calculus~\cite{wirtingermatrixcalc} and applying the complex chain rule to the cost function. The update for the complex-valued tensor becomes
\begin{align}
	\mathbf{A}_{m',n+1} &= \mathbf{A}_{m',n}  - \mu_\text{Ten}\,\frac{\partial J_n}{\partial \mathbf{A}_{m',n}^*}\\
	&= \mathbf{A}_{m',n} + 2 \mu_\text{Ten}\,e_n\, \mathbf S_{m',n}^*\,,\label{eq:ctensorupdate}
\end{align}
for all $m\in[1,M]$, with $\mathbf{S}_{m',n}$ according to \eqref{eq:stlmsrev}.
%
%
The CLMS weight update stays the same as in~\eqref{eq:CLMSupdate}. This change now fully supports the complex domain without having to repeat filters (i.e. two tensor or LMS blocks).

In terms of normalization, the first two architectures utilize the same update for $\mu_\text{Ten}$ as presented in~\cite{TensorPaper}, the normalization of the CLMS is straight forward as well and works as presented in~\eqref{eq:CLMSupdate} for both, TTLMS and CTLMS. In order to normalize the complex-valued tensor in the CTLMS architecture, the same principle as for a real-valued one is used, i.e. the error is approximated via the first order complex-valued Taylor expansion~\cite{jointNorm}  
\begin{equation}
	\label{eq:taylor_exp}
	\begin{split}
		e_{n+1} \approx e_n &+ \sum_{r = 1}^{R}\frac{\partial \tilde e_n}{\partial \mathbf A_{m'}(:,r)}\Delta \mathbf A_{m'}(:,r) \\
		&+ \sum_{r = 1}^{R}\frac{\partial \tilde e_n}{\partial \mathbf A_{m'}^*(:,r)}\Delta \mathbf A_{m'}^*(:,r)
	\end{split}
\end{equation}
where $\tilde e_n = y_n - \mathbf w_n^{\text T } \tilde{\mathbf z}_n.$. The term $\Delta \mathbf A_{m'}(:,r) = 2 e_n\, \mathbf S_{m',n}^*$ follows directly from~\eqref{eq:ctensorupdate} and
\begin{align}
	\frac{\partial \tilde{ e}_n}{\partial \mathbf A_{m'}(:,r)}& = -\sum_{p = 1}^P w_{n,p}
	\begin{pmatrix}
		\mathbf 0_{i_{m',n-p+1}-1 \times 1} \\
		\circledast_{m\neq m'} \mathbf A_{m,n-p+1}(i_{m,n-p+1},:) \\
		\mathbf 0_{I_{m'}- i_{m',n-p+1} \times 1}
	\end{pmatrix}^{\text T}   \\ 
	& = -\mathbf S_{m',n}^{\text T}
\end{align}
therefore,
\begin{align} 
	e_{n+1} & \approx e_n  - 2 \mu_\text{Ten} e_n \sum_{r = 1}^R  \mathbf S_{m',n}^{\text T}(:,r)
	\mathbf S_{m',n}^*(:,r)\\
	& = \left (1 - 2 \mu_\text{Ten}  \norm{\mathbf S_{m',n}}_{\text F}^2 \right )e_n \,.\label{eq:error_est}
\end{align}
To maintain convergence of the algorithm~\cite{Mandic2003}, the norm of the error $e_{n+1}$ has to be smaller or equal than the norm of the right side of equation \eqref{eq:error_est}. This can be achieved when
\begin{align}
	\left |1 - 2 \mu_\text{Ten}  \norm{\mathbf S_{m',n}}_{\text F}^2 \right | < 1.
\end{align}
Solving this equation for $\mu_\text{Ten}$, the normalization can be introduced by replacing $\mu_\text{Ten}$ in~\eqref{eq:ctensorupdate} by
\begin{align}
	\mu_{\text{Ten},n} = \frac{\bar{\mu}_\text{Ten}}{\epsilon + \norm{\mathbf S_{m',n}}_{\text F}^{2}}, \qquad 0 < \bar{\mu}_\text{Ten} < 1\,.
\end{align}
\section{Complexity}

\begin{table*}
	\centering
	\caption{Complexity in terms of multiplications, additions and divisions of the considered algorithms depending on the parameters of the methods for both, estimation (forward) and update (backward) steps.}
\resizebox{2\columnwidth}{!}{%
	\begin{tabular}{c|c|ccc}
		\multicolumn{2}{c|}{Algorithm}             			& Mult. & Add. & Div. \\ \hline
		
		\multirow{2}{*}{TLMS-2R}       	        & Forward  	    &  $2P+2R(M-1)$                 & $2P+2R-4$  &  --  \\
		& Backward      &  $2MR(P(M-1)+I_m)+4P(M+1)+2$  & $4P+2MRI_m(P+1)$  & $2+2M$  \\ \hline
		\multirow{2}{*}{TTCLMS}     				& Forward  	    &  $4P+2R(M-1)$                 & $4P+2R-4$  &  --  \\
		& Backward      &  $2MR(P(M-1)+I_m)+4P(M+2)+4$  & $12P+2MRI_m(P+1)+2$  & $1+2M$  \\ \hline
		\multirow{2}{*}{CTLMS}       				& Forward  	    &  $4P+4R(M-1)$                 & $6P+2R(M+1)-8$  &  --  \\
		& Backward      &  $4MR(P(M-1)+I_m)+8P(M+1)+4$  & $2MR(P(M-1)+2I_m(P+1))+4P(M+1)+2$  & $1+1M$ 
\end{tabular}}
\label{tab:compl1}
\vspace{-0.5cm}
\end{table*}

The computational complexity in terms of additions, multiplications and divisions for all architectures is depicted in Table~\ref{tab:compl1}. The complexity is the least for the first architecture, which just repeats the tensor and LMS blocks for both paths, and is highest with the fully complex-valued architecture. However, it is important to note that the fully complex implementation is able to leverage the full information present in the real and imaginary parts of all signals, while the other two architectures are not able to achieve this.

		
\section{Simulations}

To evaluate the proposed models for their performance on a complex-valued system identification example, we chose the well-known case of transmitter (Tx) induced harmonics which can occur in 4G/5G cellular transceivers in the case of downlink carrier aggregation coupled with a non-ideal Tx power amplifier (PA). For more details on the exact signal model, the reader is referred to~\cite{9404290}. Additionally, this model simulates saturation behavior of the PA which might occur if the Tx signal power is close to the limit of the PA's dynamic range~\cite{TensorPaper}. Therefore, the interference signal we want to estimate is $y_n = \mathbf{h}_\text{Dup}^\text{T} \, \mathbf{z}_n + \eta_n$, 
%
%
where $\mathbf{h}_\text{Dup}\in\mathbb{C}^P$ constitutes the complex-valued stop-band frequency response of a linear filter (the so-called duplexer), $\eta_n$ constitutes a noise term, 
\begin{equation}
	\mathbf{z}_n = \left[\frac{x_n^2}{1+\abs{x_n}}, \frac{x_{n-1}^2}{1+\abs{x_{n-1}}}, \ldots, \frac{x_{n-P}^2}{1+\abs{x_{n-P}}}\right]\,,
\end{equation}
are the complex-valued transmit samples after the PA and $x_n$ is modeled as colored noise, i.e. $x_n = a \, x_{n-1} + \sqrt{1-a^2} \nu_n$, 
%
%
with $\nu_n$ denoting complex-valued white Gaussian noise. The used evaluation metric is the mean squared error (MSE), defined as
\begin{equation}
	\text{MSE}_\text{dB} = 10\log_{10}\left(\frac{1}{{L}}\sum_{{l}=1}^{{L}}\left({d}_{n}^{(l)}-\hat{y}_{n}^{(l)}\right)^2\right)\,,
\end{equation}
where ${d}_{n}^{(l)}$ is the desired signal, $\hat{y}_{n}^{(l)}$ is the estimate at time $n$ of the $l$-th run, and $L$ is the total number of runs.

For the simulations we chose a filter order of $P=16$, the memory, i.e. dimensionality $M$, of all tensors is two (one dimension for the real- and imaginary parts of the input signal, respectively), the rank of all tensors has been chosen empirically and is set to $R=10$, and the length of the (C)LMS filters has been chosen to coincide with $P$. The step-sizes for the tensors are $\mu_\text{Ten} = 0.009$, $\mu_\text{Ten} = 0.009$, $\mu_\text{Ten} = 0.075$ and for the (C)LMS $\mu_\text{LMS} = 0.009$, $\mu_\text{LMS} = 0.005$, $\mu_\text{LMS} = 0.009$, for the first, second and third architectures shown in Fig.~\ref{fig:model}, respectively, and all regularization parameters have been set to $\epsilon=10^{-12}$. Lastly, the signal $d_n$ resides $\SI{10}{\db}$ above $\eta_n$.

\begin{figure}[t!]
	\centering
	\includegraphics[width=0.75\columnwidth]{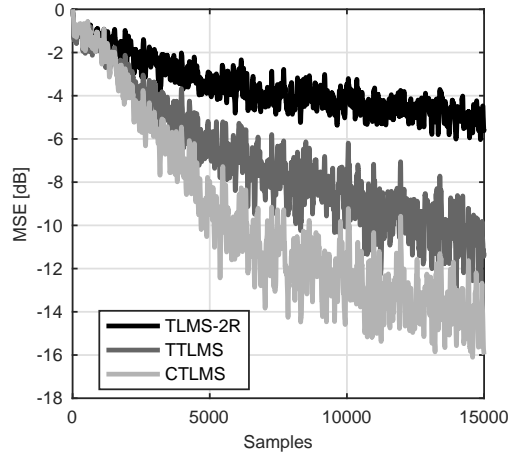}
	\caption{MSE curves for the considered scenario.}
	\label{fig:sim}
	\vspace{-0.5cm}
\end{figure}

The comparison of all three proposed architectures is shown in Fig.~\ref{fig:sim}, where the simulation was repeated and averaged over $L=20$ different real-life duplexer fittings. It can be seen that the first architecture, which just repeats the processing pipeline of the original real-valued algorithm twice, performs the worst. Using a complex-valued LMS filter with two tensors already drastically improves performance, and as expected, the fully complex-valued architecture yields the best overall performance.

\section{Conclusion}

In this paper we extended current state-of-the-art architectures for system identification via a joint tensor-LMS based framework to complex valued models. We proposed three different architectures that comply with complex-valued models. The first architecture simply repeats the estimation blocks (tensor and LMS) for the real and imaginary paths. While this is the most straight-forward approach, it yields poor performance as the two paths cannot interact with each other. To mitigate this problem for the linear subsystem, the LMS block has been replaced with a CLMS filter in our second architecture, which showed moderate improvements compared to the previous case. To fully leverage the complex valued approach, we finally proposed an architecture that models all sub-systems in a complex manner, i.e. via a complex-valued tensor and CLMS filter. This final solution significantly outperforms both other architectures in our considered application.

\bibliographystyle{IEEEtran}
\bibliography{mybib}

\end{document}